\def\year{2021}\relax
\documentclass[letterpaper]{article} 
\usepackage{aaai21}  
\usepackage{times}  
\usepackage{helvet} 
\usepackage{courier}  
\usepackage[hyphens]{url}  
\usepackage{graphicx} 
\usepackage{natbib}  
\usepackage{caption} 
\usepackage{microtype}
\usepackage{subfigure}
\usepackage{booktabs} 

\usepackage{amsmath,amsfonts,amsthm} 
\usepackage{algorithm}

\usepackage{algpseudocode}
\usepackage{enumitem}
\usepackage{multirow}
\usepackage{bbm}

\usepackage{xcolor}
\usepackage{colortbl}
\usepackage{booktabs}

\usepackage[switch]{lineno}

\newtheorem{definition}{Definition}
\newtheorem{assumption}{Assumption}

\newtheorem{theorem}{Theorem}
\newtheorem{lemma}{Lemma}

\title{Robustness of Accuracy Metric and its Inspirations in Learning with Noisy Labels}
\newcommand*\samethanks[1][\value{footnote}]{\footnotemark[#1]}
\author{
    Pengfei Chen,\textsuperscript{\rm 1}~
    Junjie Ye,\textsuperscript{\rm 2}\thanks{Corresponding authors.}~
    Guangyong Chen,\textsuperscript{\rm 3}\samethanks~
    Jingwei Zhao,\textsuperscript{\rm 2}~
    Pheng-Ann Heng\textsuperscript{\rm 1,3}\\
}
\affiliations {
    \textsuperscript{\rm 1} The Chinese University of Hong Kong\\
    \textsuperscript{\rm 2} VIVO AI Lab\\ 
    \textsuperscript{\rm 3} Guangdong Provincial Key Laboratory of Computer Vision and Virtual Reality Technology, Shenzhen Institutes of Advanced Technology, Chinese Academy of Sciences\\
    \{pfchen, pheng\}@cse.cuhk.edu.hk, \{junjie.ye, jingwei.zhao\}@vivo.com, gy.chen@siat.ac.cn
}

\begin{document}
\maketitle

\begin{abstract}
For multi-class classification under class-conditional label noise, we prove that the accuracy metric itself can be robust. We concretize this finding's inspiration in two essential aspects: training and validation, with which we address critical issues in learning with noisy labels. For training, we show that maximizing training accuracy on sufficiently many noisy samples yields an approximately optimal classifier. For validation, we prove that a noisy validation set is reliable, addressing the critical demand of model selection in scenarios like hyperparameter-tuning and early stopping. Previously, model selection using noisy validation samples has not been theoretically justified. We verify our theoretical results and additional claims with extensive experiments. We show characterizations of models trained with noisy labels, motivated by our theoretical results, and verify the utility of a noisy validation set by showing the impressive performance of a framework termed noisy best teacher and student (NTS). Our code is released~\footnote{\url{https://github.com/chenpf1025/RobustnessAccuracy}}.
\end{abstract}

\section{Introduction}
In real-world classification tasks, annotation methods such as crowdsourcing systems~\cite{yan2014learning} and online queries~\cite{schroff2010harvesting} inevitably introduce noisy labels. In learning with noisy labels, many works study the robustness, providing theoretical guarantees for some robust loss functions~\cite{ghosh2017robust,zhang2018generalized,charoenphakdee2019symmetric,xu2019l_dmi,ma2020normalized}: a classifier minimizing the loss on noisy distribution is guaranteed to minimize the loss on clean distribution. These works commonly assume that the label noise is conditional on the true class, i.e., the class-conditional noise~\cite{natarajan2013learning}.

\begin{figure*}[ht]
	\centering
	\includegraphics[width=1.7\columnwidth]{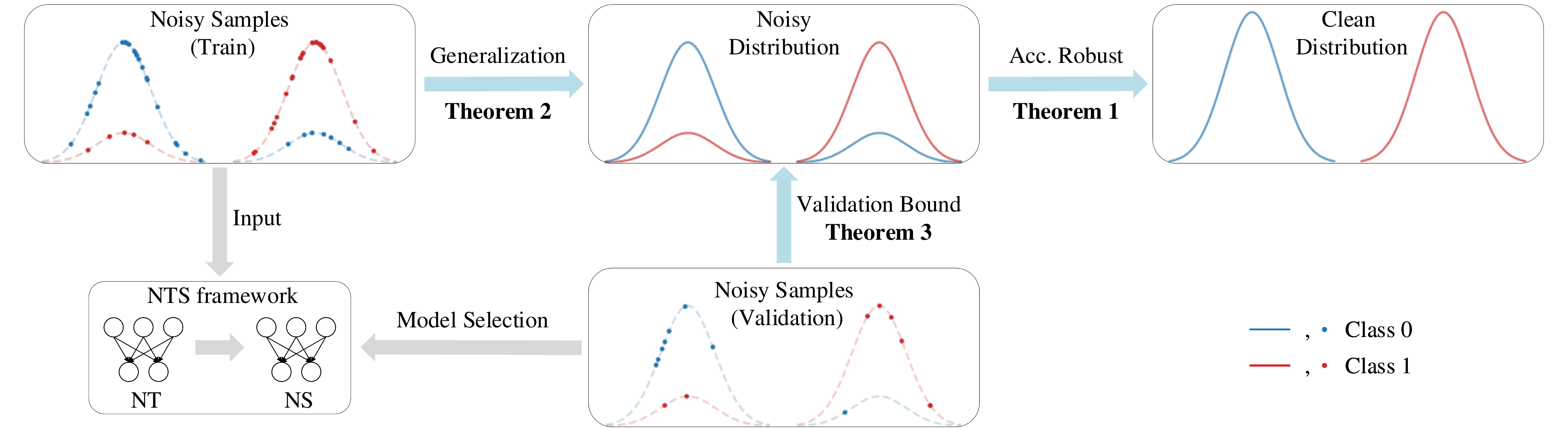}
	\caption{A systematical illustration of learning with noisy labels.}
	\label{fig_main}
\end{figure*}

In this paper, we first prove that the accuracy metric itself is robust for common diagonally-dominant class-conditional noise, i.e., \textit{a classifier maximizing its accuracy on the noisy distribution is guaranteed to maximize the accuracy on clean distribution}. It seems counterintuitive since we previously believe maximizing accuracy on noise results in overfitting. In fact, Deep Neural Networks (DNNs) can achieve $100\%$ training accuracy on finite samples~\cite{zhang2017understanding} but not the noisy distribution. Theorem~\ref{theorem_robust} shows that for any classifier, the maximized accuracy on noisy distribution is $1-\varepsilon$, and we obtain an optimal classifier if the maximum is attained, where $\varepsilon$ is the noise rate. The definition of robustness used in Theorem~\ref{theorem_robust} is consistent with existing works on robust loss functions. On the one hand, such robustness is not sufficient to guarantee `good' generalization performance when trained with finite noisy samples because by sampling, we can not directly optimizing the classifier w.r.t. a distribution. On the other hand, we can still show inspirations of Theorem~\ref{theorem_robust} by analyzing the gaps between training/validation accuracy on finite samples and the accuracy on noisy distribution.

For training, even if without any robust loss functions or advanced training strategies, \textit{we can obtain an approximately optimal classifier by maximizing training accuracy on sufficiently many noisy samples}. This claim is justified by Theorem~\ref{theorem_robust} and an additional Theorem~\ref{theorem_gen}, which presents a counterpart of the generalization bound derived from Vapnik–Chervonenkis (VC) dimension~\cite{vapnik1999overview,data2012scholarly}.  For validation, the accuracy on hold-out noisy samples is an unbiased estimator of the accuracy on noisy distribution. Therefore, \textit{a noisy validation set is reliable}. Together with Theorem~\ref{theorem_robust}, the validation bound presented in Theorem~\ref{theorem_val} formally justifies the utility of a noisy validation set. Some previous works~\cite{zhang2018generalized,nguyen2019self,xia2019anchor,xia2020part} empirically use a noisy validation set to tune hyperparameters, while our contribution is the theoretical justification.

In experiments, we focus on verifying our theoretical results and additional claims. We verify characterizations of models derived from Theorem~\ref{theorem_robust}$\&$\ref{theorem_gen}, including a dummy classifier given few samples, predicting exactly the noise transition process given more samples, and approaching a global optimal classifier with sufficiently many noisy samples. To show the utility of a noisy validation, we present a framework termed Noisy best Teacher and Student (NTS), which selects a teacher with the highest accuracy on a noisy validation set and retrains a student. It improves many baselines on CIFAR-10/100 with synthetic noise and Clothing1M~\cite{xiao2015learning} with real-world noise.

\section{Backgrounds}
For a $K$-classes classification problem, we denote the feature space by $\mathcal{X}$ and the label space by $\mathcal{Y}=\{1,2,\cdots,K\}$. Let $D_{X,Y}$ be the clean distribution of $(X,Y)\in\mathcal{X}\times\mathcal{Y}$. With label noise, we have only access to samples from a noisy distribution $\tilde{D}_{X,\tilde{Y}}$ of $(X,\tilde{Y})\in\mathcal{X}\times\mathcal{Y}$. The random variable of noisy labels $\tilde{Y}$ is corrupted from $Y$ via a noise transition matrix $T\in[0,1]^{K\times K}$, s.t., 
\begin{equation}
\mathrm{Pr}[\tilde{Y}=j|Y=i]=T_{i,j}.
\end{equation}
Then the noise rate is $\varepsilon=1-\sum_{i\in\mathcal{Y}}\mathrm{Pr}[Y=i]T_{i,i}$. Such a noise model is common in the literature~\cite{ghosh2017robust,zhang2018generalized,han2018masking,charoenphakdee2019symmetric,yao2020dual}, known as class-conditional noise~\cite{natarajan2013learning}, s.t.,
\begin{assumption}
	\label{assumption_independent}
	$X$ is independent of $\tilde{Y}$ conditioning on $Y$.
\end{assumption}

A classifier is denoted by $h\in\mathcal{H}:\mathcal{X}\mapsto\mathcal{Y}$, where $\mathcal{H}$ is the hypothesis space. The performance of a classifier is evaluated by its accuracy on clean distribution
\begin{equation}
A_D(h) :=\mathbb{E}_{(X,Y)\sim D}[\textbf{1}(h(X)=Y)] = \mathrm{Pr}[h(X)=Y],
\end{equation}
where $\textbf{1}(h(X)=Y)$ is the indicator function. By training a classifier on a collection of noisy samples $\tilde{S}=\{(x_i,\tilde{y}_i)\}_{i=1}^{m}$ drawn i.i.d. from $\tilde{D}_{X,\tilde{Y}}$, we can obtain the training accuracy as following,
\begin{equation}
A_{\tilde{S}}(h) := \frac{1}{m}\sum_{i=1}^{m}\textbf{1}(h(x_i)=\tilde{y}_i).
\end{equation}
Thus, an essential problem is to bound the gap between $A_{\tilde{S}}(h)$ and $A_D(h)$. To this end, we introduce the accuracy on noisy distribution,
\begin{equation}
A_{\tilde{D}}(h) :=\mathbb{E}_{(X,\tilde{Y})\sim\tilde{D}}[\textbf{1}(h(X)=\tilde{Y})] = \mathrm{Pr}[h(X)=\tilde{Y}].
\end{equation}
In this paper, Theorem~\ref{theorem_robust} bridges the gap between $A_{\tilde{D}}(h)$ and $A_D(h)$ by showing the robustness of the accuracy metric. With Theorem~\ref{theorem_robust}, the remaining issue is the gap between $A_{\tilde{S}}(h)$ and $A_{\tilde{D}}(h)$, as if without noise.

Following~\cite{chen2019understanding}, we use a confusion matrix $C(h)\in[0,1]^{K\times K}$ to characterize the performance of $h$ in detail,
\begin{equation}
C_{i,j}(h):=\mathrm{Pr}[h(X)=j|Y=i].
\end{equation}
\section{Related works}
\label{sec_related}
There have been many robust loss functions~\cite{ghosh2017robust,zhang2018generalized,charoenphakdee2019symmetric,xu2019l_dmi,ma2020normalized}, whereas the $0-1$ loss~\cite{bartlett2006convexity}, justified to be robust under binary classification~\cite{manwani2013noise,ghosh2015making}, has not attracted enough attention. We previously believe that to combat noisy labels, a loss function should be not only robust, but also easy to optimize in training. In this paper, we first show that accuracy metric itself is robust for multi-class classification under common diagonally-dominant class-conditional noise. We then present several related findings useful in training and validation, where the optimization is not a concern since we can easily maximize training accuracy with a surrogate loss such as the cross-entropy loss.

For the generalization bound, the Vapnik-Chervonenkis (VC) dimension~\cite{vapnik1999overview,data2012scholarly} and the Natarajan dimension~\cite{natarajan1989learning,daniely2014optimal} are classical techniques that have been used in analysing label noise~\cite{natarajan2013learning,khetan2018learning}. In this paper, thanks to Theorem~\ref{theorem_robust}, a simple counterpart of the generalization bound derived from VC dimension is sufficient to justify our claim that we can obtain an approximately optimal classifier by maximizing training accuracy on sufficiently many noisy samples. It further motivates us to present a thorough characterizations of models trained with noisy labels, including a dummy classifier given few samples, predicting exactly the noise transition process given more samples, and approaching a global optimal classifier with sufficiently many noisy samples.

For robust training methods, apart from using a robust loss function, one can adopt sample-selection/weighting~\cite{malach2017decoupling,han2018co,jiang2018mentornet,hendrycks2018using,yu2019does,fang2020rethinking}, loss-correction~\cite{reed2014training,patrini2017making}, label-correction~\cite{tanaka2018joint,zheng2020error} and other refined training strategies~\cite{nguyen2020self,li2020dividemix}. Specifically, the idea of using teacher predictions is common in loss-correction~\cite{arazo2019unsupervised}, label-correction~\cite{tanaka2018joint} and the classical distillation~\cite{hinton2015distilling}. It is known that early-stopped teachers~\cite{arpit2017closer,cho2019efficacy} may be better. Our NTS's critical point is a noisy validation set, theoretically justified by Theorem~\ref{theorem_robust}$\&$\ref{theorem_val}, with which we can easily select a good teacher. Thus, we can obtain impressive performance using merely cross-entropy loss, without explicitly preventing overfitting using advanced training techniques. In contrast, existing methods such as \citet{tanaka2018joint} uses regularization to avoid memorizing noise when training the teacher network. Our performance gain is not simply due to distilling a teacher, but due to selecting a good teacher using a noisy validation set. Otherwise, distilling a converged model, which memorizes noise, can not improve the generalization much. Moreover, our aim is not merely presenting the algorithm NTS and comparing it with baselines, but to verify the utility of a noisy validation set with NTS's impressive performance. Justifying the utility of a noisy validation set both theoretically and empirically is a critical contribution, considering that almost all methods need to tune hyperparameters, whereas the usage of a noisy validation set has been empirically demonstrated~\cite{zhang2018generalized,nguyen2019self,xia2019anchor,xia2020part} but has not been theoretically justified.
\section{Main results} 
In a classification task, our ultimate goal is obtaining a global optimal classifier that shall generalize well on clean distribution.
\begin{definition}
	\label{def_classifier_opt}
	(Global optimal classifier). The global optimal classifier, denoted by $h^{\star}$, is a classifier such that $h^{\star}(X)=Y$ almost everywhere, i.e., the accuracy on clean distribution is $A_D(h^{\star})=1$. Equivalently, the confusion matrix $C(h^{\star})=I$, where $I$ is a $K\times K$ identity matrix.
\end{definition}
The definition holds except for a trivial case where there exists uncertainty for $Y$ conditioning on fully knowing $X$.

\subsection{The accuracy metric is robust}
\label{sec_understand_opt_noise}
We first show that \textit{the accuracy metric itself is robust to diagonally-dominant noise}, where the diagonally-dominant condition is a mild assumption commonly adopted in literature~\cite{ghosh2017robust,zhang2018generalized,chen2019understanding}.
\begin{assumption}
	\label{assumption_T}
	The noise transition matrix $T$ is diagonally-dominant, i.e., $\forall i, T_{i,i}>\max_{j\in\mathcal{Y},j\neq i}T_{i,j}$.
\end{assumption}

\begin{theorem} 
	\label{theorem_robust}
	With Assumption~\ref{assumption_independent}$\&$\ref{assumption_T}, if there exists a global optimal classifier $h^{\star}\in\mathcal{H}$, then\\
	(i) $h^{\star}$ is the classifier that maximizes the accuracy on the noisy distribution, i.e.,
	\begin{equation}
	\label{eq_acc_max}
	\max_{h\in\mathcal{H}}A_{\tilde{D}}(h) = A_{\tilde{D}}(h^{\star})=1-\varepsilon,
	\end{equation}
	where $\varepsilon=1-\sum_{i\in\mathcal{Y}}\mathrm{Pr}[Y=i]T_{i,i}$ is the noise rate;
	
	(ii) when the accuracy of a classifier on noisy distribution approaches its maximum, the classifier will approach the global optimal classifier, i.e.,
	\begin{equation}
	A_{\tilde{D}}(h)\rightarrow\max_{h\in\mathcal{H}}A_{\tilde{D}}(h) \implies A_D(h)\rightarrow 1.
	\end{equation}
\end{theorem}
In Appendix~\ref{sec_app_proof}, we show a convergence rate $1-A_{D}(h)\leq \frac{1}{\min\limits_{i,j\in\mathcal{Y},j\neq i}(T_{i,i}-T_{i,j})}\cdot (\max_{h\in\mathcal{H}}A_{\tilde{D}}(h) - A_{\tilde{D}}(h))$. For any classifier, the maximum of accuracy on noisy distribution is $1-\varepsilon$ rather than $1$. It is not contradictory to the observation that we can achieve $100\%$ training accuracy on finite samples with sufficiently powerful DNNs~\cite{zhang2017understanding} because by sampling, we are maximizing the accuracy on finite samples rather than the accuracy on noisy distribution. On the one hand, we must be aware that the robustness, as in its widely used form in many existing works, is not sufficient to guarantee `good' generalization performance when trained with finite noisy samples. On the other hand, we can still show the inspirations of Theorem~\ref{theorem_robust} from two essential aspects: training and validation, by analyzing the gaps between accuracy on finite samples and the accuracy on noisy distribution.

\subsection{Training with noisy labels}
\label{sec_vc}
In this section, we analyze the generalization to noisy distribution given noisy samples drawn from it. Intuitively, the training accuracy is a biased estimator of accuracy on noisy distribution. The bias converges to $0$ with sufficiently many training samples. This intuition is formalized by the generalization bound in the following Theorem~\ref{theorem_gen}. The bound is a counterpart of the generalization bound derived from VC dimension~\cite{vapnik1999overview,data2012scholarly}. It can be generalized very well to multi-class classification by Natarajan dimension~\cite{natarajan1989learning,daniely2014optimal}. We first recall the definition of shattering and VC dimension, which can be interpreted as a measurement of the complexity/expressivity of all possible classifiers in a hypothesis space.
\begin{definition}
	(Shattering and VC dimension~\cite{vapnik1999overview,data2012scholarly}). Let $\mathcal{H}$ be the hypothesis space of all possible classifiers (functions). We say a set of $m$ points $\{x_1,x_2,\cdots,x_m\}\subset\mathcal{X}$ is \textbf{shattered} by $\mathcal{H}$ if all possible binary labeling of the points can be realized by functions in $\mathcal{H}$. The \textbf{VC dimension} of $\mathcal{H}$, denoted by $d_{VC}(\mathcal{H})$, is the cardinality of the largest set of points in $\mathcal{X}$ that can be shattered by $\mathcal{H}$.
\end{definition}
\begin{theorem}
	\label{theorem_gen}
	Considering training a classifier $h\in\mathcal{H}$ on a collection of noisy samples $\tilde{S}=\{x_i,\tilde{y}_i\}_{i=1}^{m}$ drawn i.i.d. from $\tilde{D}_{X,\tilde{Y}}$, where the hypothesis space $\mathcal{H}$ has a finite VC dimension. Then with probability at least $1-\delta$,
	\begin{scriptsize}
	\begin{equation}
	\begin{aligned}
	\label{eq_gen}
	A_{\tilde{D}}(h) - A_{\tilde{S}}(h) \geq -\sqrt{\frac{8\left(d_{VC}\cdot\left(\ln(2m/d_{VC})+1\right)+\ln(4/\delta)\right)}{m}},
	\end{aligned}
	\end{equation}
	\end{scriptsize}
	where $A_{\tilde{S}}(h)$ is the training accuracy on noisy samples and $A_{\tilde{D}}(h)$ is the accuracy on noisy distribution.
\end{theorem}

The proof is in Appendix~\ref{sec_app_proof}. The bound converges to $0$ with sufficiently large $m$. The significance is how the bound helps us understand the essence of learning with noisy labels. Firstly, the bound implies that with sufficiently many noisy samples, the accuracy on noisy distribution is approximately lower bounded by the training accuracy. In this case, if we maximize training accuracy, the accuracy on noisy distribution is approximately maximized. Then Theorem~\ref{theorem_robust} implies that the classifier approaches an optimal classifier. Therefore, we conclude that we can obtain an approximately optimal classifier by maximizing training accuracy on sufficiently many noisy samples. Moreover, the results motivate characterizations of neural networks trained with noisy labels. In particular, with respect to training size $m$, models trained with noisy labels can be typically categorized into the following three cases:
\begin{itemize}
	\item [1)] If $m$ is quite small such that the model can not learn any generalized features, then the generalization accuracy typically has a high error and a high variance.
	\item [2)] If $m$ is relatively large such that the distribution of $X$ is well approximated yet the model still fits all training samples, then $A_{\tilde{S}}(h)\approx 1$, $C(h)\approx T$ and $A_D(h)\approx 1-\varepsilon$.
	\item [3)] If $m$ is sufficiently large, then the bound in Theorem~\ref{theorem_gen} approaches $0$, and the highest training accuracy that any classifier can achieve approaches $1-\varepsilon$. When the maximum is attained, we approximately obtain a global optimal classifier, s.t., $A_{\tilde{S}}(h)\approx 1-\varepsilon$, $C(h)\approx I$ and $A_D(h)\approx 1$.
\end{itemize}
Case 1) is obvious if training samples are limited; case 2) has been demonstrated in the literature~\cite{chen2019understanding}; case 3) is derived from Theorem~\ref{theorem_robust}$\&$\ref{theorem_gen}. In practice, apart from these typical cases, we can observe transition phase, e.g., 1) $\rightarrow$ 2) and 2) $\rightarrow$ 3) as we gradually increase $m$. We will demonstrate these cases in experiments.

\begin{figure}
	\centering
	\includegraphics[width=\columnwidth]{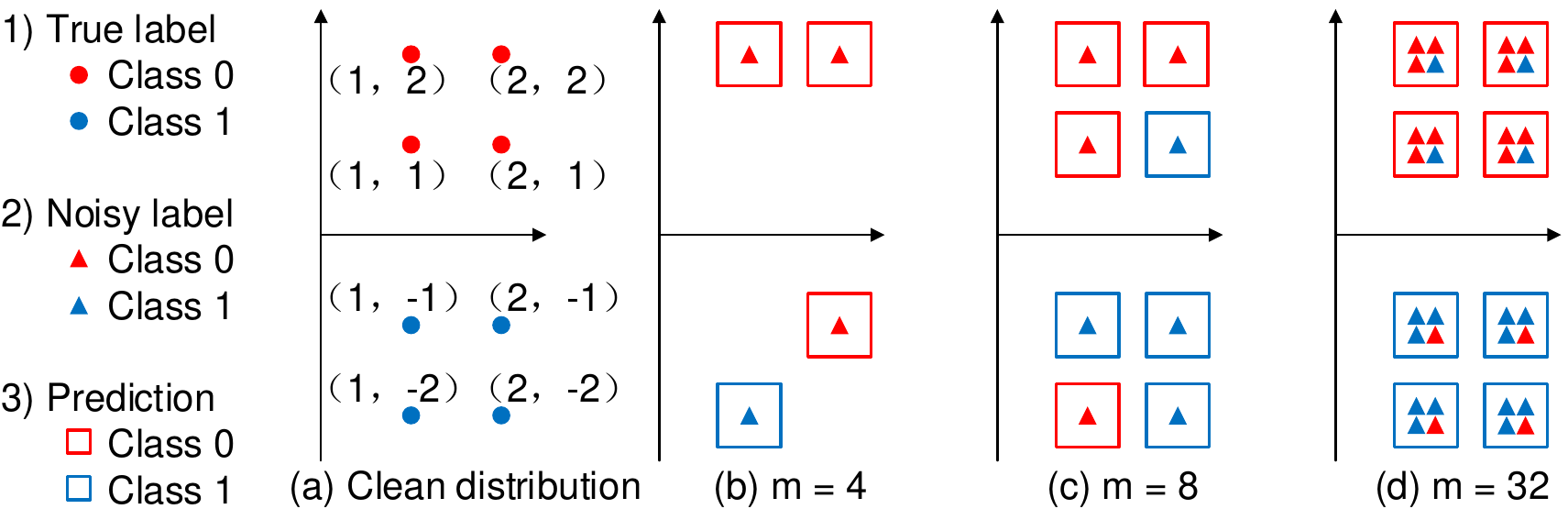}
	\caption{A tabular example of classification under noisy labels. (a): the distribution of data points with true labels. (b-d): examples of sampling $m$ instances with noisy labels for training, where $m=4, 8, 32$. In (d), we slightly shift the overlapped samples for display.}
	\label{fig_example}
\end{figure}

To quickly understand the characterizations, we present a tabular example. As shown in Figure~\ref{fig_example}~(a), we show a classification problem on a discrete space, s.t., $\mathcal{X}=\{(1,\pm2), (2,\pm2), (1,\pm1), (2,\pm1)\}$, $\mathcal{Y}=\{0,1\}$, where samples in $\{(1,2), (2,2), (1,1), (2,1)\}$ belong to class $0$ and the rest belong to class $1$. The data has a uniform distribution, s.t., $\mathrm{Pr}[X=x]=1/8,\forall x\in\mathcal{X}$. For the noisy distribution, we use a noise transition matrix $T=[0.75, 0.25;\,0.25,0.75]$, i.e, each sample has a probability of $1/4$ to be mislabeled. Now we sample m $i.i.d.$ samples for training, the example sampling results and predictions of a classifier that maximizes training accuracy are illustrated in Figure~\ref{fig_example}~(b-d).
\begin{itemize}
	\item [1)] $m=4$, Figure~\ref{fig_example}~(b). The optimal training accuracy is $1$, and the resulting classifier can not generalize well. Since the classifier's prediction on data-points without a training sample is unknown, the testing accuracy is unknown.
	\item [2)] $m=8$, Figure~\ref{fig_example}~(c). The optimal training accuracy is $1$, and the resulting classifier $h$ has a confusion $C(h)=T$. The testing accuracy on clean distribution is $0.75$, which equals to $1-\varepsilon$.
	\item [3)]  $m=32$, Figure~\ref{fig_example}~(d). The optimal training accuracy is $0.75$, which equals to our theoretical result $1-\varepsilon$. The resulting classifier is a global optimal classifier with testing accuracy $1$ on the clean distribution.
\end{itemize}

\subsection{Validation with noisy labels}
Validation is a crucial step in the standard machine learning pipeline. When there is label noise, how to conduct validation reliably without clean samples is rarely discussed in previous works. A noisy validation set is used in some previous works~\cite{zhang2018generalized,nguyen2019self,xia2019anchor,xia2020part} but without theoretical justification. In this paper, our Theorem~\ref{theorem_robust} implies that we can approach the optimal classifier by approximately maximizing the accuracy on noisy distribution. Taking a step further, we see that the accuracy on hold-out noisy samples is an unbiased estimator of the accuracy on noisy distribution. Therefore, a noisy validation set is reliable. The reliability is measured by a validation bound presented in the following Theorem~\ref{theorem_val}, which is very tight even with few noisy validation samples.
\begin{theorem}
	\label{theorem_val}
	Given a classifier $h$, considering validating it on a noisy validation set $\tilde{V}=\{(x_i,\tilde{y}_i)\}_{i=1}^{n}$ drawn i.i.d. from $\tilde{D}_{X,\tilde{Y}}$ and observing the validation accuracy
	\begin{equation}
	A_{\tilde{V}}(h) := \frac{1}{n}\sum_{i=1}^{n}\textbf{1}(h(x_i)=\tilde{y}_i),
	\end{equation} 
	then for any $0<\delta\leq1$, we have that with probability at least $1-\delta$,
	\begin{equation}
	\label{eq_val}
	A_{\tilde{D}}(h) - A_{\tilde{V}}(h) \geq -\sqrt{\dfrac{\ln(1/\delta)}{2n}},
	\end{equation}
	where $A_{\tilde{D}}(h)$ is the accuracy on noisy distribution.
\end{theorem}
The proof is in Appendix~\ref{sec_app_proof}. To see how tight the bound is, considering $n=1000$ noisy samples and $\delta=0.01$, the gap in Eq.~(\ref{eq_val}) is no larger than $0.048$. It means we shall obtain an error no larger than $0.048$ with at least probability $0.99$. Theorem~\ref{theorem_robust}$\&$\ref{theorem_val} justify that it is reliable to use a small noisy validation set for model selection. Previously, there has not been a theoretically widely-accepted criterion for validation without clean samples. For example, many previous experiments on CIFAR-10 use the $10k$ clean samples to tune hyperparameters, which may not be practical in learning with noisy labels since we assume no access to many clean samples. Therefore, our theoretical justification for a noisy validation set is critical, especially considering lack of clean samples and the unavoidable needs of model selection in scenarios such as hyperparameters-tuning and early stopping.

\section{Experiments}
\begin{figure*}[ht]
	\centering
	\includegraphics[width=1.7\columnwidth]{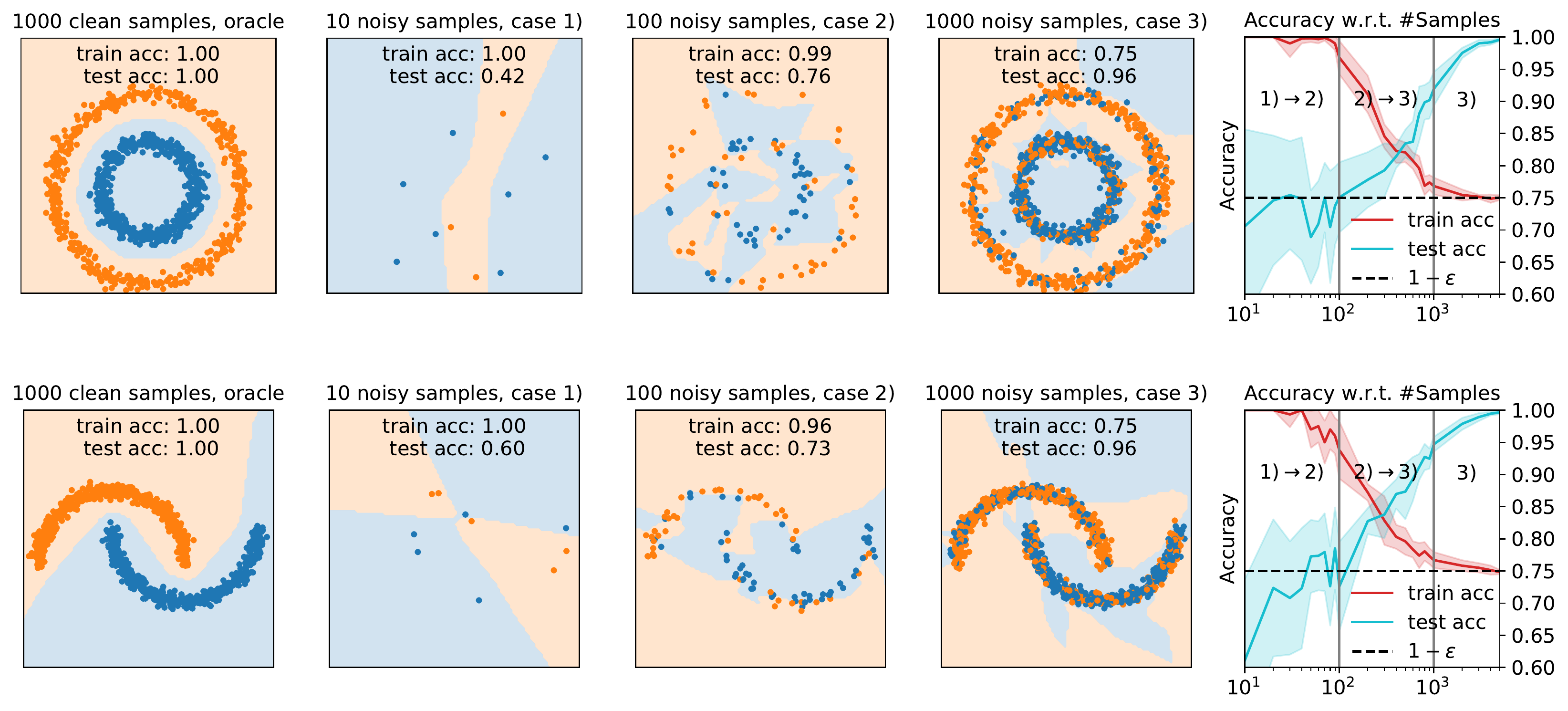}
	\vspace{-0.15in}
	\caption{Synthetic training samples and the classifier's decision regions. From left to right: an oracle case trained on clean samples; three typical cases when trained with noisy labels; train/test accuracy (at the last training step) w.r.t. number of training samples. Orange: class 0; Blue: class 1. First raw: `circles' dataset; Second raw: `moons' dataset. The $\varepsilon$ is noise rate.}
	\label{fig_synthetic}
\end{figure*}

\subsection{Training with noisy labels - characterizations}
Theorem~\ref{theorem_robust}$\&$\ref{theorem_gen} imply that we can obtain an optimal classifier by maximizing training accuracy on sufficiently many noisy samples, even if without any advanced training strategies. Models' characterizations motivated by the theoretical results have been discussed in Section~\ref{sec_vc}. Here we conduct experiments to visualize neural networks' typical characterizations when trained with noisy labels. As shown in Figure~\ref{fig_synthetic}, we adopt two classical synthetic datasets `circles' and `moons', and pollute them with a noise transition matrix $T=[0.7,0.3; 0.2,0.8]$. We implement an MLP with two hidden layer of size $32$. We train the classifier using SGD optimizer with full batch for at most $20K$ steps to ensure the training accuracy is optimized until convergence, and test on $10k$ clean samples. We vary the training sizes to demonstrate different characterizations and repeat each experiment $10$ times. Results in Figure~\ref{fig_synthetic} cover the three typical characterizations summarized in Section~\ref{sec_vc}. We also present curves of train/test accuracy at the last step w.r.t. number of training samples, where the shadow indicates the standard deviation. It clearly demonstrates transitions $1)\rightarrow2)\rightarrow3)$ as the number of training samples increases. For case 2), the training accuracy is approximately $1$, and the testing accuracy is around $1-\varepsilon=0.75$; as the number of samples increase, the trained network finally approaches case 3), with the training accuracy converging to $1-\varepsilon$, and the testing accuracy approaching $1$. For deep networks, we may not have sufficient samples to approximately obtain an optimal classifier. Still, Theorem~\ref{theorem_robust}$\&$\ref{theorem_gen} and numerical experiments here justify the utility of data augmentation techniques~\cite{cubuk2019autoaugment,he2016deep} in learning with noisy labels: data augmentation narrows the gap between noisy samples and noisy distribution. Previously, it is intuitively believed but not explained that why augmented noisy samples, which have the same noise, improves generalization.

\subsection{A noisy validation set is reliable}

To verify the utility of a noisy validation set, which has been theoretically justified by Theorem~\ref{theorem_robust}$\&$\ref{theorem_val}, we present the NTS framework. The implementation consists of two steps, (1) noisy best teacher (NT): training on given noisy labels and selecting a teacher model with the highest accuracy on a noisy validation set, (2) noisy best student (NS): training on labels predicted by NT and selecting a student model similarly. We empirically validate NTS on extensive benchmark datasets, including CIFAR-10 and CIFAR-100 with uniform noise and realistic asymmetric noise flipped between similar classes such as DEER$\rightarrow$HORSE, as well as the large-scale benchmark Clothing1M~\cite{xiao2015learning}, which contains 1 million real-world noisy training samples.

\begin{table*}[t]
	\caption{Results of WRN-28-10 on CIFAR-10 and CIFAR-100 under uniform noise and asymmetric noise. Each experiment is repeated three times, i.e., 288 trials in total. The best accuracy on each basis is in bold.}
	\label{tab_cifar}
	\begin{center}
	\begin{scriptsize}
	\begin{tabular}{lcccccccc}
		\toprule
		\multirow{2.5}{*}{Dataset} & \multirow{2.5}{*}{Basis} & \multirow{2.5}{*}{Model} & \multicolumn{3}{c}{Uniform Noise} & \multicolumn{3}{c}{Asymmetric Noise} \\ \cmidrule(lr){4-6}\cmidrule(lr){7-9}
		&							 &                         & 0.2       & 0.4       & 0.6       & 0.2        & 0.3        & 0.4        \\
		\midrule
		\multirow{14}{*}{CIFAR-10}
		& \multirow{3}{*}{CE}
		&Last		&$86.30\pm0.20$			&$74.18\pm2.27$			&$59.15\pm1.31$			&$89.95\pm1.23$			&$86.75\pm0.64$			&$79.47\pm1.60$			\\
		&&NT		&$93.63\pm0.21$			&$90.09\pm0.41$			&$82.48\pm0.47$			&$94.59\pm0.10$			&$93.33\pm0.32$			&$89.07\pm1.44$			\\
		&&NS		&$\mathbf{95.73\pm0.11}$ &$\mathbf{93.76\pm0.18}$ &$\mathbf{87.88\pm0.23}$ &$\mathbf{95.09\pm0.10}$ &$\mathbf{94.80\pm0.44}$ &$\mathbf{91.49\pm1.68}$ \\ \cmidrule(lr){2-9}
		& \multirow{3}{*}{GCE}
		&Last		&$95.02\pm0.22$			&$92.72\pm0.25$			&$86.99\pm0.24$			&$94.15\pm0.45$			&$91.55\pm0.65$			&$82.41\pm1.61$			\\
		&&NT		&$94.81\pm0.07$			&$92.53\pm0.04$			&$87.11\pm0.10$			&$94.41\pm0.17$			&$92.72\pm0.08$			&$87.60\pm1.15$			\\
		&&NS		&$\mathbf{95.34\pm0.24}$ &$\mathbf{94.50\pm0.30}$ &$\mathbf{91.32\pm0.12}$ &$\mathbf{95.08\pm0.09}$ &$\mathbf{94.11\pm0.08}$ &$\mathbf{90.43\pm0.73}$ \\ \cmidrule(lr){2-9}
		& \multirow{3}{*}{Co-T}
		&Last		&$95.09\pm0.07$			&$92.18\pm0.36$			&$87.32\pm0.38$			&$93.56\pm0.08$			&$91.39\pm0.29$			&$88.79\pm0.36$			\\
		&&NT		&$95.17\pm0.14$			&$92.42\pm0.29$			&$87.18\pm0.29$			&$94.44\pm0.14$			&$92.15\pm0.14$			&$89.93\pm0.10$			\\
		&&NS		&$\mathbf{96.26\pm0.14}$ &$\mathbf{94.73\pm0.20}$ &$\mathbf{92.11\pm0.40}$ &$\mathbf{95.47\pm0.09}$ &$\mathbf{93.73\pm0.36}$ &$\mathbf{91.32\pm0.30}$ \\ \cmidrule(lr){2-9}
		& \multirow{3}{*}{DMI}
		&Last		&$94.20\pm0.20$			&$41.60\pm29.08$		&$11.68\pm0.72$			&$95.29\pm0.21$			&$36.29\pm37.87$		&$10.35\pm2.76$			\\
		&&NT		&$\mathbf{94.25\pm0.30}$ &$91.10\pm0.46$		&$79.38\pm5.17$			&$95.14\pm0.26$			&$94.12\pm0.52$			&$89.04\pm3.31$			\\
		&&NS		&$94.10\pm0.11$			&$\mathbf{91.11\pm0.53}$ &$\mathbf{83.46\pm0.46}$ &$\mathbf{95.29\pm0.08}$ &$\mathbf{95.03\pm0.08}$ &$\mathbf{92.28\pm1.33}$			\\
		\midrule
		\multirow{14}{*}{CIFAR-100}
		& \multirow{3}{*}{CE}
		&Last		&$70.61\pm0.20$			&$55.96\pm0.62$			&$38.25\pm0.58$			&$70.10\pm0.01$			&$61.51\pm0.21$			&$51.11\pm0.13$			\\
		&&NT		&$73.38\pm0.47$			&$66.46\pm0.26$			&$57.94\pm0.30$			&$74.20\pm0.35$			&$68.27\pm0.44$			&$54.46\pm1.22$			\\
		&&NS		&$\mathbf{76.85\pm0.29}$ &$\mathbf{71.71\pm0.59}$ &$\mathbf{63.69\pm0.25}$ &$\mathbf{77.07\pm0.21}$ &$\mathbf{70.51\pm0.30}$ &$\mathbf{55.61\pm1.56}$	 \\ \cmidrule(lr){2-9}
		& \multirow{3}{*}{GCE}
		&Last		&$70.95\pm0.25$			&$55.32\pm0.35$			&$20.21\pm0.59$			&$62.59\pm0.62$			&$52.75\pm0.50$			&$41.75\pm0.69$			\\
		&&NT		&$73.54\pm0.25$			&$65.48\pm0.22$			&$\mathbf{24.22\pm0.8}8$ &$71.55\pm0.17$			&$60.30\pm0.18$			&$\mathbf{43.51\pm1.69}$			\\
		&&NS		&$\mathbf{75.41\pm0.26}$ &$\mathbf{67.87\pm0.1}3$ &$23.34\pm1.30$ &$\mathbf{73.52\pm0.4}1$ &$\mathbf{61.38\pm1.04}$ &$43.07\pm1.43$ \\ \cmidrule(lr){2-9}
		& \multirow{3}{*}{Co-T}
		&Last		&$78.24\pm0.28$			&$71.63\pm0.33$			&$64.93\pm0.37$			&$74.82\pm0.07$			&$67.72\pm0.36$			&$59.05\pm1.01$			\\
		&&NT		&$78.51\pm0.20$			&$72.33\pm0.43$			&$66.32\pm0.05$			&$75.45\pm0.24$			&$69.92\pm0.39$			&$60.22\pm0.49$			\\
		&&NS		&$\mathbf{80.02\pm0.06}$ &$\mathbf{76.16\pm0.49}$ &$\mathbf{72.30\pm0.18}$ &$\mathbf{77.68\pm0.84}$ &$\mathbf{74.02\pm1.49}$ &$\mathbf{62.66\pm0.67}$ \\ \cmidrule(lr){2-9}
		& \multirow{3}{*}{DMI}
		&Last		&$73.80\pm0.40$			&$66.80\pm0.22$			&$58.09\pm0.06$			&$74.54\pm0.20$			&$68.25\pm0.24$			&$54.69\pm0.96$			\\
		&&NT		&$73.80\pm0.41$			&$66.82\pm0.16$			&$58.35\pm0.10$			&$74.79\pm0.22$			&$68.65\pm0.08$			&$54.64\pm1.05$			\\
		&&NS		&$\mathbf{73.89\pm0.34}$ &$\mathbf{66.95\pm0.23}$ &$\mathbf{58.35\pm0.06}$ &$\mathbf{74.82\pm0.23}$ &$\mathbf{68.91\pm0.41}$ &$\mathbf{54.74\pm1.03}$			\\
		\bottomrule
	\end{tabular}
	\end{scriptsize}
	\end{center}
\end{table*}

\begin{figure*}[h!]
	\centering
	\includegraphics[width=1.7\columnwidth]{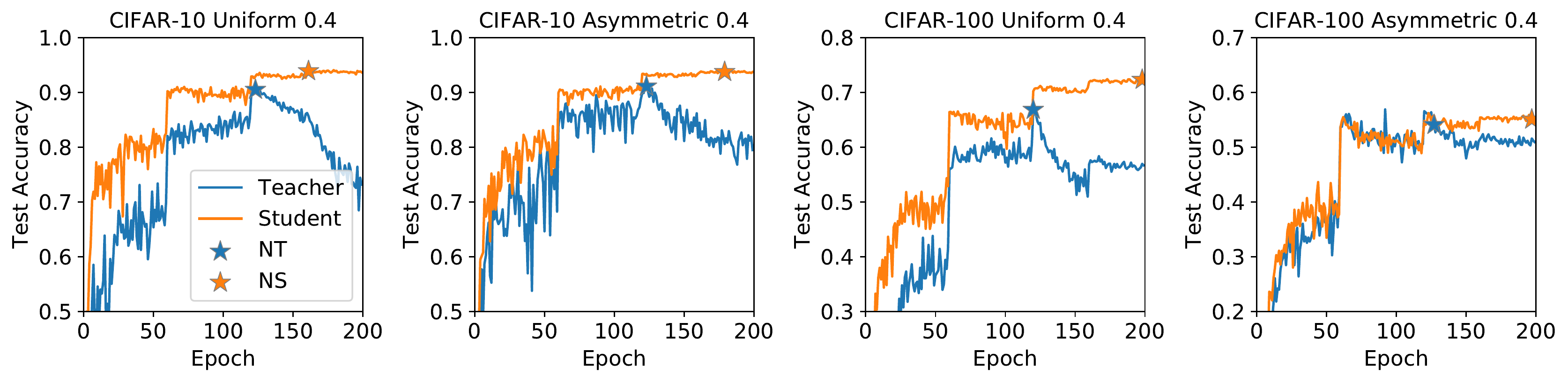}
	\vspace{-0.15in}
	\caption{Test accuracy of the teacher and student, w.r.t. training epochs. NT and NS are selected using a noisy validation set, justified by Theorem~\ref{theorem_robust} and~\ref{theorem_val}.}
	\label{fig_cifar}
\end{figure*}

\paragraph{CIFAR-10 and CIFAR-100.}

\begin{table*}[ht]
	\caption{Results of ResNet-50 on Clothing1M, using clean/noisy validation. The three results of DivideMix indicate the first model/second model/ensemble model (following \citet{li2020dividemix}). The utility of NTS and a noisy validation set is verified: 1) results of clean/noisy validation are comparable, 2) NTS obtains consistent performance gain over each baseline.}
	\label{tab_clothing}
	\begin{center}
	\begin{footnotesize}
	\begin{tabular}{cccccccccc}
		\toprule
		Basis      &\multicolumn{3}{c}{CE}    &\multicolumn{3}{c}{DMI} &\multicolumn{3}{c}{DivideMix}	\\ \cmidrule(lr){2-4}\cmidrule(lr){5-7}\cmidrule(lr){8-10}
		Validation &Last    &NT &NS &Last &NT &NS &Last &NT &NS  \\\midrule
		Clean      &$67.91$ &$68.99$ &$\textbf{70.02}$ &$71.82$ &$72.23$ &$\textbf{72.99}$ &$73.09/73.48/73.81$ &$73.98/74.31/74.65$ &$\textbf{74.48/74.70/74.89}$	\\
		Noisy      &$68.05$ &$69.12$ &$\textbf{70.21}$ &$72.01$ &$72.18$ &$\textbf{72.82}$ &$73.63/73.38/73.84$ &$73.75/73.73/74.18$ &$\textbf{73.99/74.09/74.36}$	\\
		\bottomrule
	\end{tabular}
	\end{footnotesize}
	\end{center}
\end{table*}

We use Wide ResNet-28-10 (WRN-28-10)~\cite{zagoruyko2016wide} as the classifier on CIFAR-10 and CIFAR-100. We corrupt the training set which has $50000$ samples and randomly split $5000$ noisy samples for validation. We conduct experiments on \textit{uniform noise} (symmetric noise) and \textit{asymmetric noise}, following previous settings~\cite{han2018co,patrini2017making,ren2018learning,nguyen2019self,chen2019meta}. Uniform noise is generated by uniformly flipping labels to other classes~\cite{han2018co,ren2018learning,han2020sigua}, which is the most studied scenario in literature. For asymmetric noise, following~\cite{patrini2017making,zhang2018generalized,nguyen2019self}, on CIFAR-10, it is generated by flipping labels between similar classes, i.e., TRUCK$\rightarrow$AUTOMOBILE, BIRD$\rightarrow$AIRPLANE, DEER$\rightarrow$HORSE, and CAT$\leftrightarrow$DOG, with a given probability; on CIFAR-100, it is generated by flipping each class into the next circularly with a given probability, which is also known as pair noise~\cite{han2018co}. Apart from implementing NTS on the most common Cross-Entropy (CE) loss, we show NTS is also applicable to other advanced training methods, including Generalized Cross Entropy (GCE)~\cite{zhang2018generalized} loss, which is proved to be noise-robust; Co-teaching (Co-T)~\cite{han2018co}, which is an effective method that uses two networks select small-loss training samples for each other; Determinant based Mutual Information (DMI), which is an information-theoretic robust loss. In NTS,  all student networks share the same training schedule and hyperparameters with their teachers. Strong data augmentation~\cite{cubuk2019autoaugment} is implemented, and we will show in the ablation that the improvement obtained by NTS is consistent with/without data augmentation. More training details can be found in Appendix~\ref{sec_exp_app}.

\begin{table*}[ht]
	\caption{Ablation results of WRN-28-10 trained with CE. The best accuracy is in bold.}
	\label{tab_ablation}
	\begin{center}
	\begin{footnotesize}
	\begin{tabular}{llcccccc}
		\toprule
		\multirow{2.5}{*}{Dataset} & \multirow{2.5}{*}{Model} & \multicolumn{2}{c}{No Augmentation} & \multicolumn{2}{c}{Standard Augmentation} & \multicolumn{2}{c}{Strong Augmentation} \\ \cmidrule(lr){3-4}\cmidrule(lr){5-6}\cmidrule(lr){7-8}
		 						   & 						  &Uniform 0.4       &Asymmetric 0.4      &Uniform 0.4       &Asymmetric 0.4       &Uniform 0.4       &Asymmetric 0.4               \\
		\midrule
		\multirow{3}{*}{CIFAR-10}
		&Last	&$60.94$			&$73.82$			&$70.10$			&$77.54$			&$74.18$			&$79.47$ \\
		&NT		&$72.15$			&$75.30$			&$86.14$			&$87.66$			&$90.09$			&$89.07$ \\
		&NS		&$\textbf{75.04}$ &$\textbf{76.69}$ &$\textbf{89.45}$ &$\textbf{90.23}$ &$\textbf{93.76}$ &$\textbf{91.49}$ \\
		\midrule
		\multirow{3}{*}{CIFAR-100}
		&Last	&$33.83$			&$42.31$			&$53.72$			&$49.26$			&$55.96$			&$51.11$ \\
		&NT		&$41.40$			&$41.20$			&$62.96$			&$55.92$			&$66.46$			&$54.46$ \\
		&NS		&$\textbf{41.72}$ &$\textbf{52.61}$ &$\textbf{65.92}$ &$\textbf{59.27}$ &$\textbf{71.71}$ &$\textbf{55.61}$ \\
		\bottomrule
	\end{tabular}
	\end{footnotesize}
	\end{center}
\end{table*}

Table~\ref{tab_cifar} summarizes results on CIFAR-10 and CIFAR-100, where we repeat each experiment three times. The results verify that our NTS framework and the noisy validation set not only improve performance for CE, but are also applicable to many advanced training algorithms. Firstly, in all noise settings, the best accuracy is always achieved by our NS. Secondly, a noisy validation, which is justified by Theorem~\ref{theorem_robust}$\&$\ref{theorem_val}, is an enhancement for all methods. When severe overfitting happens at the later stage (e.g., CE), we can get a significantly better NT than the last epoch. Even if existing methods can reduce overfitting, our NT still achieves higher or comparable accuracy than the last epoch. Moreover, we can not tune hyperparameters of existing methods without the noisy validation, whereas conduct validation on clean samples may violate the fair setting in learning with noisy labels. Hence our Theorem~\ref{theorem_robust}$\&$\ref{theorem_val} and the NTS framework are critical, which justify the utility of a noisy validation set theoretically and empirically. Finally, we see that in most cases, we get a better NS by learning knowledge form NT. Figure~\ref{fig_cifar} shows that we benefit from early-stopped teachers~\cite{cho2019efficacy}. It is known that early stopping may help due to the memorization effect~\cite{arpit2017closer,yao2020searching}, s.t., DNNs learn simple and correct patterns first before memorizing noise. Our critical point is justifying the noisy validation set, with which we can easily select the good teachers.

\paragraph{Clothing1M.}
Clothing1M is a large-scale benchmark containing real-world noise. Strictly following the standard setting~\cite{patrini2017making,tanaka2018joint,xu2019l_dmi,li2020dividemix}, we use a ResNet-50 pre-trained on ImageNet, 1 million noisy samples for training, $14k$ and $10k$ clean data respectively for validation and test. Moreover, to demonstrate the utility of a noisy validation set, we additionally show a more challenging setting where the clean validation set is assumed unavailable. In this case, we randomly draw $14k$ samples from the noisy training set for validation. More training details can be found in Appendix~\ref{sec_exp_app}. Table~\ref{tab_clothing} summarizes results on Clothing1M. Since DivideMix~\cite{li2020dividemix} trains two networks, apart from the ensemble result obtained by averaging two networks' predictions following the original paper~\cite{li2020dividemix}, we also report results of each network for a fair comparison with other baselines. For NTS, when we use DivideMix as the teacher, a student is trained for each teacher, and results of each student and their average ensemble are reported similarly. In Table~\ref{tab_clothing}, we report results of the last epoch, NT and NS for each baseline method. The utility of NTS and a noisy validation set is verified on this real-world noisy dataset, because 1) results of clean/noisy validation are comparable, 2) NTS obtains consistent performance gain over each baseline. Notably, without our theoretical justification for a noisy validation set, it would be intractable to tune hyper-parameters for all baselines.

\paragraph{Ablation.}
\label{sec_ablation}
To comprehensively verify the utility of a noisy validation set when there is no augmentation, standard augmentation~\cite{he2016deep} or strong augmentation~\cite{cubuk2019autoaugment}, we conduct an ablation study on CIFAR-10 and CIFAR-100. Results are presented in Table~\ref{tab_ablation}. It is not surprising that increasing data augmentation can improve test accuracy. Still, it has not been made clear why augmented noisy samples, which have the same noisy labels, improve generalization. This paper reveals that the reason is the robustness of accuracy, which fills the gap between the noisy and clean distribution. Augmented samples, though have the same noise, can narrow the gap between training samples and the noisy distribution. In rare cases such as CIFAR-100 with $0.4$ asymmetric noise, strong augmentation can be harmful since too much randomness may result in underfitting. Importantly, we see that the performance gain of NTS, which uses a noisy validation set, is consistent under different augmentations or no augmentation. The results verify that NTS is practical and a noisy validation set is reliable.

\section{Conclusion}
In this paper, we target at revealing the essence of learning with noisy labels. We prove that the accuracy metric itself is robust for multi-class classification under common diagonal-dominant class-conditional noise, i.e., a classifier maximizing its accuracy on the noisy distribution is guaranteed to maximize the accuracy on clean distribution. We then show inspirations of this finding in two essential aspects: training and validation. For training, maximizing training accuracy on sufficiently many noisy samples yields an approximately optimal classifier. Characterizations of models trained with noisy samples are derived. For validation, it is reliable to use a noisy validation set. Justifying the noisy validation set is a critical contribution, considering that almost all methods need to tune hyperparameters, whereas conducting model-selection reliably without clean samples has not been theoretically justified in previous works. In experiments, we verify our theoretical results and additional claims by visualizing models' characterizations and demonstrating NTS's impressive performance.

\clearpage

\section{Acknowledgments}
The work is supported by the Key-Area Research and Development Program of Guangdong Province, China (2020B010165004) and the National Natural Science Foundation of China (Grant Nos.: 62006219, U1813204).

\bibliography{paper}

\clearpage
\appendix
\onecolumn

\section{Proofs}
\label{sec_app_proof}
\begin{lemma}
	\label{lemma_independent}
	With Assumption~\ref{assumption_independent}, given any classifier $h:\mathcal{X}\mapsto\mathcal{Y}$, $\tilde{Y}$ is independent of $h(X)$ conditioning on $Y$.
\end{lemma}
The conclusion easily follows from Shannon's information theory~\cite{brillouin2013science} since $\tilde{Y}$ is independent of $X$ conditioning on $Y$.

\begin{proof} (\textbf{Theorem~\ref{theorem_robust}}). 
(i) Without loss of generality, we assume each true class has nonzero probability, i.e., $\mathrm{Pr}[Y=i]>0,\forall i\in\mathcal{Y}$. Otherwise we can remove the class with probability $0$.

Recall that the confusion matrix $C(h)$ for a classifier h is defined as $C_{i,j}(h):=\mathrm{Pr}[h(X)=j|Y=i]$. Thus, $0\leq C_{i,j}(h)\leq 1$ and $\sum_{j\in\mathcal{Y}}C_{i,j}(h)=1$. 

With Assumption~\ref{assumption_T}, we have, $\forall$ $C(h)$ and $i\in\mathcal{Y}$,
$\sum_{j\in\mathcal{Y}}T_{i,j}\cdot C_{i,j}(h)\leq \sum_{j\in\mathcal{Y}}T_{i,i}\cdot C_{i,j}(h)=T_{i,i}.$ 

With Lemma~\ref{lemma_independent}, we have, $\forall i,j\in\mathcal{Y}$,
$\mathrm{Pr}[h(X)=j|Y=i,\tilde{Y}=j]=\mathrm{Pr}[h(X)=j|Y=i]=C_{i,j}(h).$
Therefore,
\begin{equation}
\label{eq_noisy_acc_app}
\begin{aligned}
A_{\tilde{D}}(h)\quad&=\quad\mathrm{Pr}[h(X)=\tilde{Y}]
\quad=\quad\sum_{i\in\mathcal{Y}}\mathrm{Pr}[Y=i]\cdot\mathrm{Pr}[h(X)=\tilde{Y}|Y=i]\\
&=\quad\sum_{i,j\in\mathcal{Y}}\mathrm{Pr}[Y=i]\cdot\mathrm{Pr}[\tilde{Y}=j|Y=i]\cdot\mathrm{Pr}[h(X)=j|Y=i,\tilde{Y}=j]\\
&=\quad\sum_{i\in\mathcal{Y}}(\mathrm{Pr}[Y=i]\sum_{j\in\mathcal{Y}}T_{i,j}\cdot C_{i,j}(h))
\quad\leq\quad\sum_{i\in\mathcal{Y}}\mathrm{Pr}[Y=i]\cdot T_{i,i}
\quad=\quad1-\varepsilon.
\end{aligned}
\end{equation}
The maximum is attained if and only if $C(h)=I$, i.e., when $h$ is a global optimal classifier $h^{\star}$.

(ii) Note that $\sum_{j\in\mathcal{Y}}C_{i,j}(h)=1,\forall i\in\mathcal{Y}$. Using Eq.~(\ref{eq_noisy_acc_app}), we obtain the gap between $A_{\tilde{D}}(h)$ and its maximum,
\begin{equation}
\label{eq_noisy_acc_gap_app}
\begin{aligned}
&\max_{h\in\mathcal{H}}A_{\tilde{D}}(h) - A_{\tilde{D}}(h)
\quad=\quad\sum_{i\in\mathcal{Y}}\mathrm{Pr}[Y=i]\cdot T_{i,i} - \sum_{i\in\mathcal{Y}}(\mathrm{Pr}[Y=i]\sum_{j\in\mathcal{Y}}T_{i,j}\cdot C_{i,j}(h))\\
=&\sum_{i\in\mathcal{Y}}(\mathrm{Pr}[Y=i](T_{i,i}-\sum_{j\in\mathcal{Y}}T_{i,j}\cdot C_{i,j}(h)))
\quad=\quad\sum_{i\in\mathcal{Y}}(\mathrm{Pr}[Y=i](T_{i,i}\cdot\sum_{j\in\mathcal{Y}} C_{i,j}(h)-\sum_{j\in\mathcal{Y}}T_{i,j}\cdot C_{i,j}(h)))\\
=&\sum_{i,j\in\mathcal{Y},j\neq i}\mathrm{Pr}[Y=i]\cdot(T_{i,i}-T_{i,j})\cdot C_{i,j}(h).
\end{aligned}
\end{equation}

Then the conclusion follows from the bound:
\begin{equation}
0 \leq 1-A_{D}(h)\leq \frac{1}{\min\limits_{i,j\in\mathcal{Y},j\neq i}(T_{i,i}-T_{i,j})}\cdot (\max_{h\in\mathcal{H}}A_{\tilde{D}}(h) - A_{\tilde{D}}(h)).
\end{equation}

The above bound follows from Eq.~(\ref{eq_noisy_acc_gap_app}),
\begin{equation}
\begin{aligned}
&\frac{\max_{h\in\mathcal{H}}A_{\tilde{D}}(h) - A_{\tilde{D}}(h)}{\min\limits_{i,j\in\mathcal{Y},j\neq i}(T_{i,i}-T_{i,j})}
\quad=\quad\frac{1}{\min\limits_{i,j\in\mathcal{Y},j\neq i}(T_{i,i}-T_{i,j})}\cdot \sum_{i,j\in\mathcal{Y},j\neq i}\mathrm{Pr}[Y=i]\cdot(T_{i,i}-T_{i,j})\cdot C_{i,j}(h)\\
\geq&\sum_{i,j\in\mathcal{Y},j\neq i}\mathrm{Pr}[Y=i]\cdot C_{i,j}(h)
\quad=\quad\sum_{i\in\mathcal{Y}}\mathrm{Pr}[Y=i](1-C_{i,i}(h))
\quad=\quad1-\sum_{i\in\mathcal{Y}}\mathrm{Pr}[Y=i]C_{i,i}(h)\\
=&1-\sum_{i\in\mathcal{Y}}\mathrm{Pr}[Y=i]\mathrm{Pr}[h(X)=Y|Y=i]
\quad=\quad1-A_{D}(h).
\end{aligned}
\end{equation}
\end{proof}

\begin{proof} (\textbf{Theorem~\ref{theorem_gen}}). 
Note that accuracy metric is equivalent to error metric defined by $0-1$ loss. Therefore, using the uniform error bound for general $\mathcal{H}$~\cite{vapnik1999overview,data2012scholarly}, we have, $\forall\epsilon>0$,
\begin{equation}
\mathrm{Pr}\left[\sup_{h\in\mathcal{H}}\left(A_{\tilde{S}}(h)-A_{\tilde{D}}(h)\right)\geq\epsilon\right] \leq 4(2em/d_{VC}(\mathcal{H}))^{d_{VC}(\mathcal{H})}e^{-m\epsilon^2/8}
\end{equation}

Let $\delta=4(2em/d_{VC}(\mathcal{H}))^{d_{VC}(\mathcal{H})}e^{-m\epsilon^2/8}$, we get
$\mathrm{Pr}\left[\sup\limits_{h\in\mathcal{H}}\left(A_{\tilde{S}}(h)-A_{\tilde{D}}(h)\right)\geq\epsilon\right] \leq \delta.$

Thus, 
$\mathrm{Pr}\left[\inf\limits_{h\in\mathcal{H}}\left(A_{\tilde{D}}(h)-A_{\tilde{S}}(h)\right)\leq-\epsilon\right] \leq \delta.$

Then the conclusion follows by substituting
$\epsilon=\sqrt{\dfrac{8\left(d_{VC}(\mathcal{H})\cdot\left(\ln(2m/d_{VC}(\mathcal{H}))+1\right)+\ln(4/\delta)\right)}{m}}.$
\end{proof}

\begin{proof} (\textbf{Theorem~\ref{theorem_val}}). 
	The noisy validation samples $\tilde{V}=\{(x_i,\tilde{y}_i)\}_{i=1}^{n}$ are drawn i.i.d. from $\tilde{D}_{X,\tilde{Y}}$. Therefore, the indication function on each sample $\textbf{1}(h(x_i)=\tilde{y}_i), i=1,2,\cdots,n$ are i.i.d., such that $\mathbb{E}_{(X,\tilde{Y})\sim\tilde{D}}[\textbf{1}(h(X)=\tilde{Y})]=A_{\tilde{D}}(h)$. Using Hoeffding’s inequality~\cite{bentkus2004hoeffding} for these  i.i.d. Bernoulli random variables with expectation $A_{\tilde{D}}(h)$, we have
	\begin{equation}
	\mathrm{Pr}[A_{\tilde{V}}(h)-A_{\tilde{D}}(h)\geq\epsilon]\leq e^{-2n\epsilon^2},
	\end{equation}
	where $A_{\tilde{V}}(h):= \frac{1}{n}\sum_{i=1}^{n}\textbf{1}(h(x_i)=\tilde{y}_i)$ is the observed validation accuracy.
	Equivalently,
	\begin{equation}
	\mathrm{Pr}[A_{\tilde{D}}(h)-A_{\tilde{V}}(h)\leq-\epsilon]\leq e^{-2n\epsilon^2}.
	\end{equation}
	Let $\delta=e^{-2n\epsilon^2}$, then the conclusion follows by substituting $\epsilon=\sqrt{\dfrac{\ln(1/\delta)}{2n}}$.
\end{proof}

\section{More details on experiments}
\label{sec_exp_app}

Here we present more experimental details. All models are trained with Tesla V100 GPU. Our code is released~\footnote{\url{https://github.com/chenpf1025/RobustnessAccuracy}}.

\subsection{CIFAR-10 and CIFAR-100}
We add label noise to the initial training set with $50000$ samples and randomly split $5000$ noisy samples for validation. Wide ResNet-28-10 (WRN-28-10)~\cite{zagoruyko2016wide} is used as the classifier. Our NTS framework is implemented on four typical methods: Cross Entropy (CE), Generalized Cross Entropy (GCE)~\cite{zhang2018generalized}, Co-teaching (Co-T)~\cite{han2018co} and Determinant based Mutual Information (DMI). Though each basic method may require some specific hyperparameters, we ensure that in each method, training of the teacher and the student share exactly the same setting.

CE, GCE and Co-T shares the same batch size of 128 and learning rate schedule, i.e., training using SGD optimizer for 200 epochs, with a initial learning rate of 0.1, which is decreased by a factor of 5 after 60, 120 and 160 epochs. Following its original paper and official implementation, DMI uses a model pretrained by CE as initialization and requires a larger batch size of 256 and a smaller learning rate, which is tuned in $\{10^{-4}, 10^{-5}, 10^{-6}\}$ and fixed to $10^{-6}$ finally. It is trained using SGD optimizer for 100 epochs without learning rate change. In all methods, the SGD optimizer is implemented with momentum 0.9 and weight decay $5\times10^{-4}$. For GCE, following its original paper, there is a warm-up epoch, before which no loss pruning is applied. It is set as the epoch of first and second learning rate change for CIFAR-10 and CIFAR-100 respectively~\cite{zhang2018generalized}. We tune the warm-up epoch in $\{30, 60, 120\}$ and finally use $60$, $120$ for CIFAR-10 and CIFAR-100 respectively. For Co-T, we also find that a warm-up (tuned in $\{30, 60, 120\}$) where no sample selection is applied can help the training, especially under strong noise, hence we use a warm-up epoch $60$ for $0.6$ uniform noise and $0.4$ asymmetric noise.

We use strong augmentation and dropout $0.2$ for all methods by default, except for GCE on CIFAR-100, where we found it works better with standard augmentation and no dropout. Here, standard augmentation~\cite{he2016deep} includes per-pixel normalization, horizontal random flip and $32\times32$ random crop after padding with 4 pixels on each side. Strong augmentation~\cite{cubuk2019autoaugment} includes operations described in~\cite{cubuk2019autoaugment}, i.e., ShearX/Y, TranslateX/Y, Rotate, AutoContrast, Invert, Equalize, Solarize, Posterize, Contrast, Color, Brightness, Sharpness and Cutout~\cite{devries2017improved}. In the ablation study presented in Section~\ref{sec_ablation}, we conduct experiments without augmentation, with standard augmentation and with strong augmentation, which verifies that the utility of NTS and a noisy validation set is consistent in all these settings.

\subsection{Clothing1M}
Strictly following the standard setting~\cite{patrini2017making,tanaka2018joint,xu2019l_dmi,li2020dividemix}, we use a ResNet-50 pre-trained on ImageNet, 1 million noisy samples for training, $14k$ and $10k$ clean data respectively for validation and test. For DMI~\cite{xu2019l_dmi} and DivideMix~\cite{li2020dividemix}, we reproduce results following their official implementations. In NTS, we find that simply training the student using normal cross-entropy loss works well. We train the ResNet-50 using SGD optimizer for 10 epochs, with a batchsize 256 and an initial learning rate of 0.001, which is decreased by a factor of 10 after 5 epochs. The SGD optimizer is implemented with momentum 0.9 and weight decay 0.001. We use standard data augmentation with per-pixel normalization, horizontal random flip and $224\times224$ random crop. Note that DivideMix~\cite{li2020dividemix} trains two networks, hence we train a student for each teacher. 

Using exactly the same training hyperparameters, we also demonstrate a more challenging setting, where the $14k$ clean validation samples are assumed unavailable. We randomly sample $14k$ noisy samples (i.e., $1k$ samples each class) from the 1 million noisy training samples for validation. This setting further verifies the utility of a noisy validation set.

\end{document}